\pdfoutput=1

\documentclass[11pt]{article}

\usepackage{acl}

\usepackage{times}
\usepackage{latexsym}
\usepackage{amsmath}
\usepackage{float}  %
\usepackage{placeins}  %
\usepackage[linesnumbered,ruled,vlined]{algorithm2e}
\usepackage{multirow}

\usepackage[T1]{fontenc}

\usepackage[utf8]{inputenc}
\usepackage{enumitem}

\usepackage{microtype}

\usepackage{inconsolata}

\usepackage{graphicx}
\usepackage{booktabs}  %
\usepackage{tabularx}
\usepackage{xcolor}
\usepackage[most]{tcolorbox}

\definecolor{darkgreen}{RGB}{0,100,0}
\definecolor{prompttitlebg}{RGB}{60,55,55}      %
\definecolor{promptbodybg}{RGB}{240,240,240}    %

\newtcolorbox{promptbox}[1]{
  enhanced, breakable,
  colback=promptbodybg, colframe=prompttitlebg,
  coltitle=white, fonttitle=\small\scshape\bfseries,
  title=#1,
  attach boxed title to top left={xshift=0pt, yshift=0pt},
  boxed title style={colback=prompttitlebg, sharp corners, frame hidden},
  before upper={\strut}, after upper={\strut},
  fontupper=\ttfamily\footnotesize,
  arc=0pt, outer arc=0pt, boxrule=0.5pt,
  left=6pt, right=6pt, top=6pt, bottom=6pt,
}

\usepackage[compact]{titlesec}
\titlespacing{\section}{0pt}{6pt plus 2pt minus 2pt}{3pt plus 1pt minus 1pt}
\titlespacing{\subsection}{0pt}{5pt plus 2pt minus 1pt}{2pt plus 1pt minus 1pt}
\titlespacing{\subsubsection}{0pt}{4pt plus 1pt minus 1pt}{1pt plus 1pt minus 1pt}
\titlespacing{\paragraph}{0pt}{3pt plus 1pt minus 1pt}{1em}

\setlist[itemize]{topsep=2pt, partopsep=0pt, itemsep=1pt, parsep=0pt}
\setlist[enumerate]{topsep=2pt, partopsep=0pt, itemsep=1pt, parsep=0pt}

\AtBeginDocument{%
  \setlength{\abovedisplayskip}{6pt plus 2pt minus 2pt}
  \setlength{\belowdisplayskip}{6pt plus 2pt minus 2pt}
  \setlength{\abovedisplayshortskip}{3pt plus 1pt minus 1pt}
  \setlength{\belowdisplayshortskip}{3pt plus 1pt minus 1pt}
}

\newcommand{\system}{EviMem}
\newcommand{\memmod}{LaceMem}
\newcommand{\retmod}{IRIS}

\title{EviMem: Evidence-Gap-Driven Iterative Retrieval for Long-Term Conversational Memory}

\author{Yuyang Li$^{1*}$\quad
Yime He$^{2*}$\quad
Zeyu Zhang$^{2*}$\quad
Dong Gong$^{2\dag}$\\
$^1$The Australian National University\quad
$^2$UNSW Sydney\\
\small $^*$Equal contribution. $^\dag$Corresponding author: dong.gong@unsw.edu.au.}

\begin{document}
\maketitle
\begin{abstract}
Long-term conversational memory requires retrieving evidence scattered across multiple sessions, yet single-pass retrieval fails on temporal and multi-hop questions. Existing iterative methods refine queries via generated content or document-level signals, but none explicitly diagnoses the \emph{evidence gap}---what is missing from the accumulated retrieval set---leaving query refinement untargeted. We present \textit{\system{}}, combining \textit{\retmod{}} (Iterative Retrieval via Insufficiency Signals), a closed-loop framework that detects evidence gaps through sufficiency evaluation, diagnoses what is missing, and drives targeted query refinement, with \textit{\memmod{}} (Layered Architecture for Conversational Evidence Memory), a coarse-to-fine memory hierarchy supporting fine-grained gap diagnosis. On LoCoMo, \system{} improves Judge Accuracy over MIRIX on temporal (73.3\%$\to$81.6\%) and multi-hop (65.9\%$\to$85.2\%) questions at 4.5$\times$ lower latency.
Code:~\url{https://github.com/AIGeeksGroup/EviMem}.
\end{abstract}

\newcommand{\zz}[1]{\textcolor{red}{[Zeyu: #1]}}

\section{Introduction}
\label{sec:intro}

LLM-based agents increasingly rely on long-term memory to reason over interaction histories spanning multiple sessions \citep{Park2023GenerativeAgents, zhong2023memorybank}.
While direct lookups, \emph{``What is Bob's favorite restaurant?''}, can be resolved via single-pass retrieval, questions requiring temporal reasoning (\emph{``When did Alice change her career plans?''}) or multi-hop synthesis (\emph{``What dance piece did Jon's team perform to win first place?''}) demand evidence scattered across conversations that share no overlapping keywords.
For such questions, single-pass retrieval systematically fails to surface all relevant facts.

This failure reflects a well-documented architectural limitation: conventional retrieval pipelines operate as open-loop, single-pass sequences with no feedback between stages~\citep{hu2025memory}.
Queries are constructed before any evidence is examined and cannot adapt to retrieval outcomes \citep{pereira2023visconde, gao2023precise}.
Semantic drift and forced top-$K$ selection introduce noise that compounds for multi-hop and temporal questions.

Several recent methods address this by introducing iterative retrieval.
IRCoT \citep{trivedi2023ircot} uses each reasoning step as a re-query.
Iter-RetGen \citep{shao2023iterretgen} uses the model's generated output as context for subsequent retrieval.
FLARE \citep{jiang2023flare} triggers re-retrieval when generation confidence drops below a threshold.
Self-RAG \citep{asai2024selfrag} emits reflection tokens to score individual passages.
CRAG \citep{yan2024crag} classifies individual documents as correct, ambiguous, or incorrect to trigger corrective retrieval.
Yet their sufficiency signals operate at a finer granularity than the retrieval set as a whole: FLARE evaluates per-token generation confidence; Self-RAG and CRAG assess individual passage quality via reflection/evaluator tokens; IRCoT validates reasoning steps; Iter-RetGen uses the intermediate generation as the next query without explicit sufficiency evaluation; ReAct-style agents decide next actions from local observations. None of these explicitly assess whether the accumulated evidence set, considered as a whole and in a tiered form (\textsc{exact} / \textsc{inferrable} / \textsc{partial}), is sufficient to answer the question. \system{} targets this collection-level, tiered sufficiency gap directly.
Without a holistic model of evidence sufficiency, these systems cannot produce targeted follow-up queries addressing diagnosed gaps; they refine blindly.
This limitation is amplified in long-term conversational memory, where evidence is temporally structured, entity-dense, and distributed across sessions with no surface similarity---a setting these methods were not designed for.

We introduce \textit{\system{}}, a conversational memory system that closes the retrieval loop via explicit \emph{evidence-gap diagnosis}---identifying what specific evidence is missing from the accumulated set, operationalized through a sufficiency evaluation mechanism.
\system{} comprises two mutually enabling components.
The retrieval component, \textit{\retmod{}} (Iterative Retrieval via Insufficiency Signals), centers on sufficiency evaluation as its primary operation.
After each retrieval iteration, an LLM evaluates the \emph{accumulated evidence set as a whole}, classifying it into three tiers—EXACT, INFERRABLE, or PARTIAL—and producing a natural-language diagnosis of missing information.
This diagnosis—rather than retrieved content or a draft answer—drives targeted query refinement.
Dual-path retrieval (one anchored to the original question, one using the refined query) prevents semantic drift, while per-entity fact tracking detects sparse entity coverage.
If evidence remains insufficient after all iterations, \retmod{} explicitly abstains instead of hallucinating.We validate the classifier's reliability, commitment precision, and confidence discriminability in \S\ref{sec:sufficiency_validation}.

The memory component, \textit{\memmod{}} (Layered Architecture for Conversational Evidence Memory), provides the substrate enabling \retmod{}’s sufficiency evaluation.
A compact \emph{Index layer} of semantic tuples supports fast search; a sparse \emph{Edge layer} enables multi-hop expansion; and a \emph{Raw layer} preserves full conversational detail for grounded generation.
This coarse-to-fine hierarchy balances structured rigidity and summarization loss~\citep{hu2025memory}. Crucially, \retmod{} relies on \memmod{}’s atomic tuples for fine-grained sufficiency evaluation, while \memmod{} alone cannot recover distributed multi-session evidence—making the two components mutually enabling.

We evaluate on the LoCoMo benchmark \citep{maharana2024evaluating}, a long-term conversational memory dataset spanning five categories. Our main findings are:

\begin{enumerate}[leftmargin=*]
\item \textit{Evidence-gap-driven iteration disproportionately benefits complex questions.} \system{} achieves 81.6\% Judge Accuracy on temporal reasoning (from 73.3\% for the multi-agent baseline MIRIX \citep{mirix} and 58.8\% for single-pass retrieval) and 85.2\% on multi-hop questions (from 65.9\% and 81.4\%), confirming that explicit evidence-gap diagnosis is most valuable where single-pass retrieval is most likely to fail.

\item \textit{Coarse-to-fine memory enables effective sufficiency evaluation.} \memmod{}'s atomic tuple structure allows \retmod{} to diagnose evidence completeness at the right granularity. Ablations confirm incremental contributions from each component: the iterative loop, dual-layer sufficiency classification, temporal specialization, and entity tracking.

\item \textit{Competitive accuracy at a fraction of the latency.} \system{} matches or exceeds main competitor, MIRIX, accuracy at 9.54s average latency versus 42.71s, a 4.5$\times$ reduction, via replacing multi-agent orchestration with a focused iterative loop that allocates computation adaptively by question difficulty.
\end{enumerate}

\section{Related Work}

\paragraph{Long-term Conversational Memory.}
Standard RAG~\citep{lewis2021RAG} treats memory as a static collection with flat semantic matching, which is insufficient for temporal continuity across sessions.  
Persistent memory systems such as MemoryBank~\citep{zhong2023memorybank} and Generative Agents~\citep{Park2023GenerativeAgents} support long-term consistency, while recent work adds structural priors: A-Mem~\citep{xu2025mem} for agent-controlled memory selection, MAGMA~\citep{jiang2026magma} for multi-graph reasoning, and TReMu~\citep{ge2025tremu} for neuro-symbolic temporal reasoning in multi-session dialogue.
LiCoMemory and its CogniGraph component~\citep{huang2025licomemory} extend this direction with hierarchical indexing and temporal weighting, enabling more fine-grained fact organisation and time-aware retrieval across sessions. \memmod{} shares this structural philosophy but pairs it with \retmod{}'s iterative evidence-gap loop: rather than issuing a single query against the organised memory, \retmod{} evaluates whether the \emph{accumulated} evidence suffices and diagnoses exactly what remains missing, driving targeted follow-up retrieval.
These structured approaches increase expressiveness but incur higher inference overhead from multi-step graph traversal~\citep{hu2025memory}.  
\memmod{} preserves relational structure with a coarse-to-fine hierarchy—compact tuples, graph edges, and raw dialogue—while favoring parallel retrieval over sequential traversal.

\paragraph{Iterative and Adaptive Retrieval.}
Several methods move beyond single-pass RAG, differing in what drives iteration.
\emph{Content-driven} approaches use generated output as the signal: IRCoT~\citep{trivedi2023ircot} uses chain-of-thought steps, Iter-RetGen~\citep{shao2023iterretgen} uses draft answers, and FLARE~\citep{jiang2023flare} uses generation confidence.
\emph{Evaluation-driven} approaches assess retrieval quality: Self-RAG~\citep{asai2024selfrag} evaluates individual passage relevance via reflection tokens, and CRAG~\citep{yan2024crag} classifies individual documents to trigger corrective retrieval.
However, all evaluate either generated content or individual documents---none assesses whether the \emph{accumulated evidence set} suffices to answer the question.
\retmod{} fills this gap: it evaluates accumulated evidence \emph{as a whole} for sufficiency and produces an explicit diagnosis of what information remains missing, driving targeted query refinement.

\paragraph{Structured Memory and Experience Storage.}
ReasoningBank~\citep{reasoningbank} highlights the value of structured experience storage via memory-aware test-time scaling.  
HippoRAG~\citep{gutierrez2024hipporag} uses single-step PageRank over knowledge graphs, and GraphRAG~\citep{edge2024graphrag} pre-indexes community summaries for query-focused summarization; neither includes an iterative evidence-gap loop.  
Knowledge graph systems~\citep{rasmussen2025} capture entity relations but struggle with temporal nuance, while flat vector stores~\citep{packer2023, chhikara2025} lack compositional structure.
MIRIX~\citep{mirix} organizes memory into six typed components managed by specialized agents and serves as our primary baseline (\S\ref{sec:results}).

\section{Methodology}
\label{sec:method}

\subsection{\memmod{}: Layered Memory Architecture}
\label{sec:memory}

\begin{figure}[t]
  \includegraphics[width=\columnwidth]{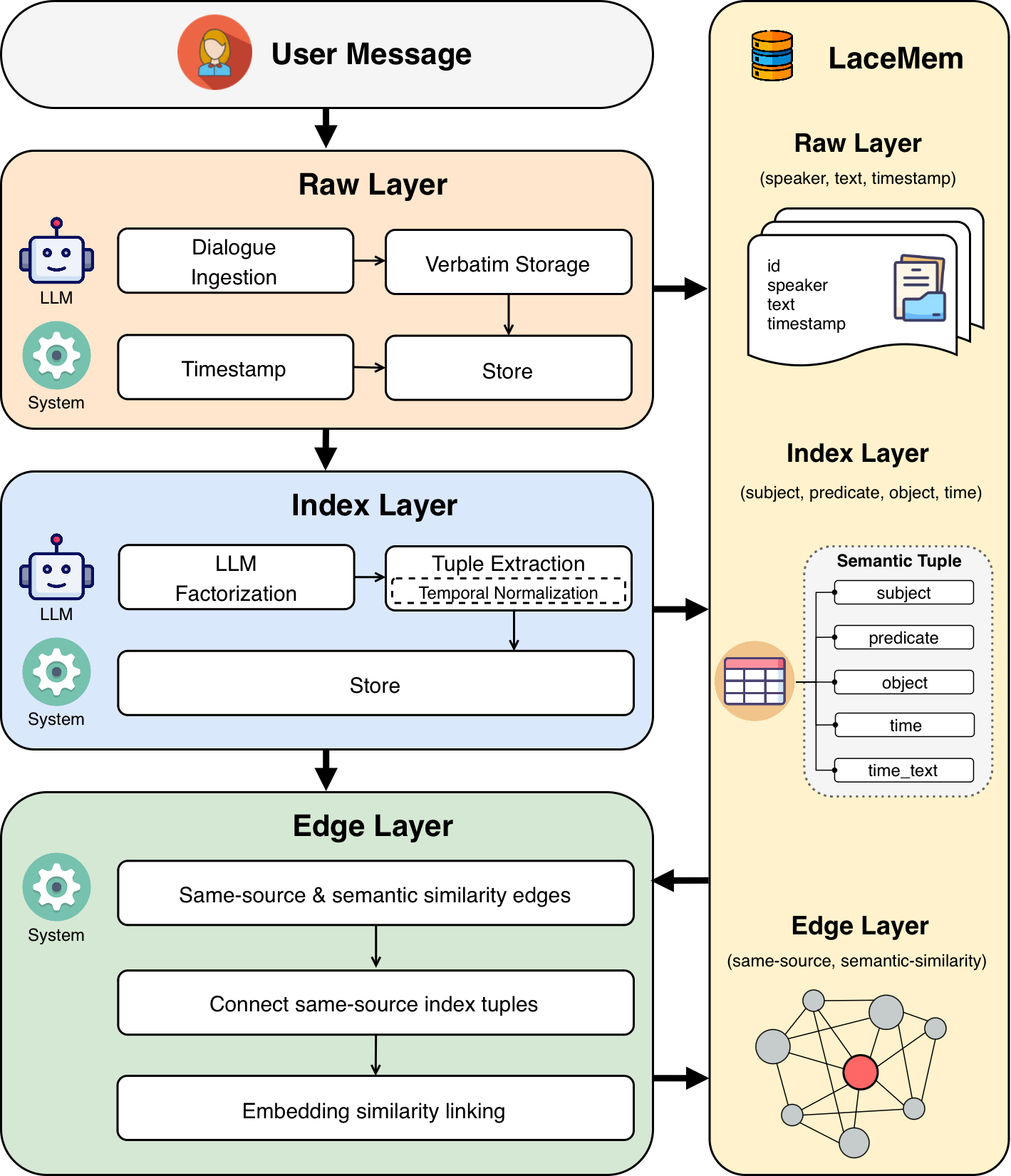}
  \caption{\memmod{} memory architecture. Dialogue is organized into three layers: Index (semantic tuples for search), Edge (graph links for multi-hop expansion), and Raw (verbatim dialogue for grounding).}
  \label{fig:raw_index_edge}
\end{figure}

Long-term conversational memory must balance \emph{fast search} over large dialogue histories, \emph{associative expansion} for multi-hop connections, and \emph{faithful grounding} in verbatim dialogue.
\memmod{} addresses this with a three-layer coarse-to-fine hierarchy (Figure~\ref{fig:raw_index_edge}): a compact \emph{Index layer} for search, a sparse \emph{Edge layer} for expansion, and a \emph{Raw layer} for grounding.
Retrieval proceeds top-down: search over Index tuples, expansion via Edge connections, and detailed grounding from the Raw layer when needed.

\paragraph{Index Layer (Search).}

Each speaker turn is factorized by an LLM into atomic semantic tuples \texttt{(subject, predicate, object, event\_time)}, with temporal expressions normalized to calendar dates. Multiple tuples may arise from one turn, and subjective statements are normalized into speaker-grounded relations (e.g., ``X is beautiful'' $\rightarrow$ ``speaker thinks X is beautiful''). Each tuple links to its source turn, enabling retrieval of full context during generation.

This atomic representation is essential for evidence-gap detection, allowing \retmod{}’s sufficiency evaluation to identify missing individual facts, e.g, session summaries or document chunks not provided.

\paragraph{Edge Layer (Expansion).}
Index tuples are linked into a sparse graph via two complementary edge types:
(1)~\textit{same-source edges} connect tuples from the same dialogue turn, preserving local coherence; and
(2)~\textit{semantic similarity edges} link related tuples across turns based on embedding similarity.
This graph enables multi-hop expansion from initially retrieved tuples, supporting associative recall across sessions that share no surface-level similarity.

\paragraph{Raw Layer (Grounding).}
Each speaker turn is stored verbatim with speaker identity and timestamp, without summarization or rewriting. Once \retmod{} retrieves relevant Index tuples and expands via Edge connections, the linked raw records provide full conversational context for grounded answer generation.

\paragraph{Co-design with \retmod{}.}
\memmod{} and \retmod{} are co-designed: each \memmod{} layer addresses a specific need of the evidence-gap loop, and together they enable capabilities neither could provide alone. The Index layer's atomic tuples give tier-level sufficiency evaluation the granularity required to detect which specific facts are missing, a precision that session summaries (GraphRAG) or passage chunks (HippoRAG) cannot match. The Edge layer turns sufficiency-identified seeds into multi-hop expansions without a separate query planner. This tight integration between memory structure and retrieval logic distinguishes \memmod{} from retrieval-method-agnostic memory systems.

\begin{figure*}[t]
\centering
\includegraphics[width=\textwidth]{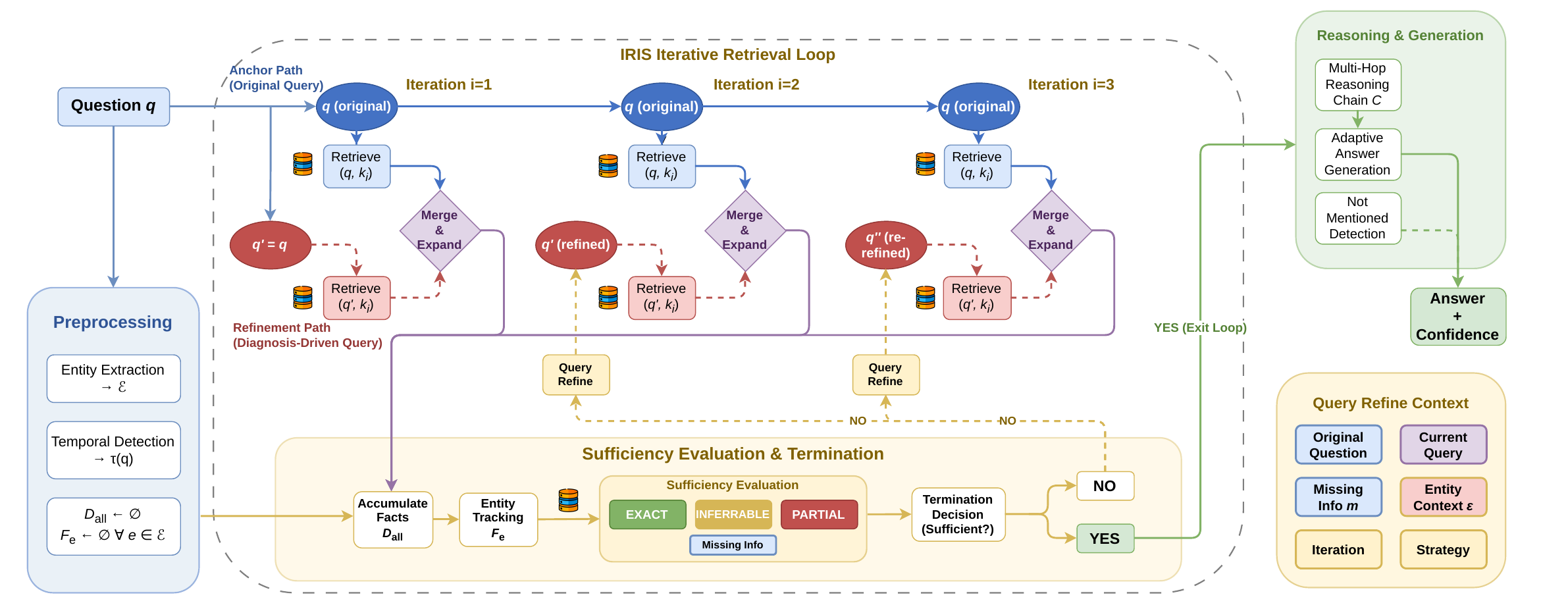}
\caption{EviMem with \retmod{} iterative retrieval pipeline. At each iteration, dual-path retrieval gathers evidence, sufficiency evaluation diagnoses gaps, and query refinement targets missing information. Retrieve in both paths and sufficiency evaluation operations are based on LaceMem.}
\label{fig:irr_overview}
\end{figure*}

\subsection{\retmod{}: Iterative Retrieval via Insufficiency Signals}
\label{sec:iris}

\retmod{} transforms the conventional open-loop retrieval into a closed-loop process driven by \emph{evidence-gap detection}: at each iteration, it evaluates whether accumulated evidence is sufficient and, if not, diagnoses what is missing.
After each retrieval iteration, an LLM classifies accumulated evidence into three tiers---EXACT, INFERRABLE, or PARTIAL---and produces a natural-language diagnosis of what information remains missing.
This diagnosis guides targeted query refinement for the next iteration. The loop continues until sufficient evidence is gathered or the maximum iteration limit is reached, after which the system either generates an answer or abstains.
Algorithm~\ref{alg:irr_loop} outlines the iterative loop; the full specification with all thresholds and constants is in Algorithm~\ref{alg:irr_loop_detailed}.

\begin{algorithm}[t]
\small
\SetAlgoVlined
\SetAlgoSkip{}
\caption{\retmod{} Iterative Retrieval Loop (see Appendix~\ref{app:algo} for the full version)}
\label{alg:irr_loop}
\KwIn{Question $q$, Memory $\mathcal{M}$, Entity set $\mathcal{E}$, Max iterations $k$}
\KwOut{Evidence $\mathcal{D}_{all}$, Confidence $c$, Evidence tier $t$}
$q' \gets q$; $\mathcal{D}_{all} \gets \emptyset$; $\mathcal{F}_e \gets \emptyset\;\forall\, e \in \mathcal{E}$\;
\For{$i \gets 1$ \KwTo $k$}{
    \tcp{Dual-path retrieval: }
    $\mathcal{D}_{anc} \gets \texttt{Retrieve}(q,\; k_i)$ \tcp*{Anchor: original question, preserves coverage}
    $\mathcal{D}_{ref} \gets \texttt{Retrieve}(q',\; k_i)$ \tcp*{Refinement: diagnosis-driven query, targets gaps}
    $\mathcal{D}^{(i)} \gets \texttt{Dedup}(\mathcal{D}_{anc} \cup \mathcal{D}_{ref} \cup \texttt{GraphExpand}(\mathcal{D}_{anc} \cup \mathcal{D}_{ref}))$\;
    $\mathcal{D}_{all} \gets \mathcal{D}_{all} \cup \mathcal{D}^{(i)}$ \tcp*{Accumulate evidence across iterations}
    \BlankLine
    \tcp{Track per-entity fact coverage for under-represented entities}
    \ForEach{$e \in \mathcal{E}$}{$\mathcal{F}_e \gets \mathcal{F}_e \cup \{d \in \mathcal{D}^{(i)} \mid e \in d\}$}
    \BlankLine
    \tcp{Evaluate accumulated evidence as a whole; $m$ diagnoses what is missing}
    $(t,\, c,\, m) \gets \texttt{EvalSufficiency}(\mathcal{D}_{all},\, q)$\;
    Calibrate $c$ based on $t$, $\tau(q)$, and entity coverage $\{|\mathcal{F}_e|\}$\;
    \BlankLine
    \tcp{Stop when evidence is sufficient or budget exhausted}
    \lIf{$\texttt{Sufficient}(t, c)$ \textbf{or} $i = k$}{\textbf{break}}
    \BlankLine
    \tcp{Use missing-information diagnosis $m$ to refine the next query}
    $q' \gets \texttt{LLM\_Refine}(q,\; q',\; m,\; \texttt{Strategy}(\tau(q), i),\; \mathcal{F}_\mathcal{E})$\;
}
\Return{$\mathcal{D}_{all},\; c,\; t$}\;
\end{algorithm}

\subsubsection{Preprocessing}
\label{sec:preprocessing}

Before entering the loop, we extract named entities from question $q$ (via pattern matching and LLM-based extraction) to obtain entity set $\mathcal{E}$, and detect temporal intent $\tau(q)$ via rule-based keyword matching.
Temporal questions trigger stricter confidence thresholds and specialized answer generation (\S\ref{sec:answer_gen}).

\subsubsection{Iterative Retrieval Loop}
\label{sec:iterative_loop}

Each iteration executes five phases, described as follows.

\paragraph{Dual-Path Retrieval.}
To balance \textit{coverage} and \textit{focus}, each iteration retrieves along two parallel paths: the \textit{anchor path} uses the original question $q$ to maintain global coverage, while the \textit{refinement path} uses the diagnosis-driven query $q_i'$ for gap-targeted search (in iteration~1, $q_1' = q$).
Retrieved facts from both paths are merged, deduplicated, and expanded via one-hop graph traversal over \memmod{}'s Edge layer.
The retrieval budget grows progressively across iterations to widen search scope when early attempts prove insufficient.

\paragraph{Entity Tracking.}
For each entity $e \in \mathcal{E}$, a fact buffer $\mathcal{F}_e$ accumulates all retrieved facts mentioning $e$ across iterations.
This per-entity view detects a failure mode invisible to query-level evaluation: in multi-entity questions, aggregate evidence may appear sufficient even though facts about a particular entity are critically sparse.

\paragraph{Sufficiency Evaluation.}
The central operation of \retmod{}: an LLM evaluates whether accumulated evidence $\mathcal{D}_{all}$ suffices to answer $q$, producing a \textit{three-tier classification}:
\begin{itemize}[nosep]
    \item \textit{EXACT}: a precise, direct answer exists in the evidence.
    \item \textit{INFERRABLE}: sufficient clues exist for reasonable inference.
    \item \textit{PARTIAL}: related evidence is present but insufficient.
\end{itemize}
Along with the tier, the evaluator produces a confidence score $c \in [0,1]$ and a natural-language description $m$ of what information remains missing.
The INFERRABLE tier bridges strict exactness and necessary inferential leaps, critical for temporal and multi-hop questions where complete evidence is rare.

\paragraph{Confidence calibration design.}
The three temporal confidence adjustments (EXACT floor 0.85, INFERRABLE cap 0.75, PARTIAL cap 0.50) are structural anchors relative to \retmod{}'s convergence thresholds. The EXACT floor equals the temporal convergence threshold ($\tau_\text{temporal} = 0.85$), guaranteeing that verbatim evidence always triggers termination regardless of the LLM's raw confidence. The INFERRABLE cap sits just above the inferrable threshold ($\tau_\text{inf} = 0.70$), permitting termination on inferred answers but not forcing it. The PARTIAL cap falls strictly below both thresholds, ensuring partial evidence never terminates the loop. Each value implements a specific convergence guarantee: EXACT always stops, INFERRABLE may stop, PARTIAL never stops. Entity tracking adds a safeguard: if any entity lacks sufficient supporting facts, the system downgrades the tier and caps confidence to prevent premature termination.

\paragraph{Termination.}
The loop terminates when sufficiency reaches EXACT or INFERRABLE with confidence above a threshold, or when the maximum iteration count $k$ is reached.

\paragraph{Query Refinement.}
When evidence gaps remain, \retmod{} generates a refined query for the next iteration.
Unlike conventional query rewriting that operates solely on the original question, \retmod{}'s refinement is \emph{diagnostic}: it is driven by the missing-information description $m$ from the sufficiency evaluator, which specifies \emph{what} the system still needs rather than blindly rephrasing.

The refinement prompt anchors on the original question to prevent drift, incorporates the gap diagnosis $m$, applies a question-type–specific search strategy, and injects entity hints when they are underrepresented. Strategy selection is rule-based; only the final query generation uses the LLM.

\subsubsection{Answer Generation}
\label{sec:answer_gen}

\paragraph{``Not Mentioned'' Detection.}
If evidence remains absent after all iterations—either no tier is activated or confidence stays extremely low—\retmod{} abstains instead of generating a hallucinated answer, converting silent failure into faithful refusal.

\paragraph{Multi-Hop Reasoning Chain.}
When evidence is classified as INFERRABLE or PARTIAL, the system generates an explicit reasoning chain that decomposes the question into sequential inference steps (e.g. \emph{``Identify Jon’s team $\to$ Find competition result $\to$ Find dance piece performed’’}) to guide answer generation.

\paragraph{Tier-Adaptive Answer Generation.}
\label{sec:tier_adaptive}
The final answer is generated by an LLM conditioned on the question, accumulated evidence, and the reasoning chain (if available). The prompt adapts to the evidence tier: with EXACT evidence the model answers directly; with inferred evidence it may qualify its response. For temporal questions with EXACT evidence, precise date/time formats are enforced. This tier-adaptive prompting avoids unnecessary hedging while allowing appropriate qualification when evidence is partial.

\subsection{Complexity Analysis}

\retmod{} incurs between $2$ and $3k{+}3$ LLM calls per question (with $k{=}3$ maximum iterations).  
In the best case, evidence is sufficient at iteration~1, requiring only one sufficiency check and one answer generation.  
In the worst case, each iteration adds a sufficiency check and query refinement, plus optional reasoning construction and final generation.  
Sufficiency checks and refinements use a lightweight LLM; only the final answer generation uses a full-scale model.  
Dual-path retrieval performs two embedding lookups per iteration, for at most $2k$ retrieval operations.

\section{Experiments}
\label{sec:results}

\subsection{Setup}

\begin{table*}[!t]
\centering
\footnotesize
\setlength{\tabcolsep}{4pt}
\begin{tabular}{llccccc}
\toprule
\textbf{Category} & \textbf{System} & \textbf{G-EVAL} & \textbf{Judge Acc} & \textbf{F1} & \textbf{ROUGE-L} & \textbf{BERTScore} \\
\midrule
\textbf{Overall} & MIRIX & 2.75 & 75.9\% & 0.113 & 0.108 & 0.840 \\
& Single-pass & 2.24 & 66.4\% & 0.127 & 0.121 & 0.844 \\
& \textbf{\system{} (Ours)} & \textbf{2.81} & \textbf{76.5}\% & \textbf{0.177} & \textbf{0.170} & \textbf{0.852} \\
\cmidrule(lr){2-7}
& \quad vs.\ MIRIX & \textcolor{darkgreen}{+2.2\%} & \textcolor{darkgreen}{+0.8\%} & \textcolor{darkgreen}{+56.6\%} & \textcolor{darkgreen}{+57.4\%} & \textcolor{darkgreen}{+1.4\%} \\
& \quad vs.\ Single-pass & \textcolor{darkgreen}{+25.4\%} & \textcolor{darkgreen}{+15.2\%} & \textcolor{darkgreen}{+39.4\%} & \textcolor{darkgreen}{+40.5\%} & \textcolor{darkgreen}{+0.9\%} \\
\midrule
\textbf{Single-hop} & MIRIX & 2.93 & 69.6\% & 0.130 & 0.112 & 0.843 \\
& Single-pass & 2.51 & 61.3\% & 0.173 & 0.155 & 0.851 \\
& \textbf{\system{} (Ours)} & \textbf{2.98} & \textbf{68.2}\% & \textbf{0.205} & \textbf{0.187} & \textbf{0.857} \\
\midrule
\textbf{Multi-hop} & MIRIX & 2.46 & 65.9\% & 0.099 & 0.091 & 0.832 \\
& Single-pass & 2.67 & 81.4\% & 0.151 & 0.138 & 0.838 \\
& \textbf{\system{} (Ours)} & \textbf{2.89} & \textbf{85.2}\% & \textbf{0.260} & \textbf{0.239} & \textbf{0.857} \\
\midrule
\textbf{Temporal} & MIRIX & 2.69 & 73.3\% & 0.082 & 0.073 & 0.833 \\
& Single-pass & 1.87 & 58.8\% & 0.069 & 0.061 & 0.839 \\
& \textbf{\system{} (Ours)} & \textbf{3.08} & \textbf{81.6}\% & \textbf{0.135} & \textbf{0.124} & \textbf{0.845} \\
\midrule
\textbf{Open-domain} & MIRIX & 3.24 & 91.6\% & 0.132 & 0.132 & 0.846 \\
& Single-pass & 2.60 & 75.5\% & 0.164 & 0.162 & 0.852 \\
& \textbf{\system{} (Ours)} & \textbf{3.17} & \textbf{85.9}\% & \textbf{0.192} & \textbf{0.191} & \textbf{0.857} \\
\midrule
\textbf{Adversarial} & MIRIX & 1.92 & 57.9\% & 0.083 & 0.082 & 0.835 \\
& Single-pass & 1.17 & 43.4\% & 0.023 & 0.024 & 0.830 \\
& \textbf{\system{} (Ours)} & \textbf{1.94} & \textbf{55.1}\% & \textbf{0.081} & \textbf{0.083} & \textbf{0.840} \\
\bottomrule
\end{tabular}
\caption{Performance across all five LoCoMo categories. G-EVAL ranges from 1--5 (higher is better).}
\label{tab:results}
\end{table*}

\paragraph{Benchmark.}
We evaluate on the LoCoMo benchmark~\citep{maharana2024evaluating}, the standard testbed for long-term conversational memory.
LoCoMo comprises 10 multi-session conversations (average 588.2 turns, 16{,}618.1 tokens) with 1{,}986 question--answer pairs spanning five categories:
(1)~\textit{Single-hop} --- answerable from a single session;
(2)~\textit{Multi-hop} --- requiring synthesis across sessions;
(3)~\textit{Temporal} --- involving time-related reasoning;
(4)~\textit{Open-domain} --- requiring external knowledge; and
(5)~\textit{Adversarial} --- unanswerable questions testing hallucination avoidance.
Gold-standard answers with evidence dialog IDs are provided for all categories.

\paragraph{Metrics.}
We report \textit{G-EVAL}~\citep{liu2023geval} (1--5 Likert scale via GPT-4 chain-of-thought) and \textit{Judge Accuracy} as primary correctness metrics, supplemented by token-level \textit{F1}, \textit{ROUGE-L}~\citep{lin2004rouge}, and \textit{BERTScore}~\citep{zhang2019bertscore}.

\paragraph{Baselines.}
We compare against two systems:
\textbf{(1)~Single-pass} uses the same \memmod{} memory (\S\ref{sec:memory}) with one-pass retrieval and direct answer generation, isolating the contribution of \retmod{}'s iterative loop.
\textit{(2)~MIRIX}~\citep{mirix} is a multi-agent memory system that decomposes memory into six typed components (Episodic, Semantic, Procedural, etc.) managed by a Meta Memory Manager via LLM tool-calling; we use it as our primary external baseline.

\paragraph{Implementation Details.}

All experiments use GPT-4o for answer generation and GPT-4o-mini for entity extraction, sufficiency evaluation, and query refinement. Retrieval uses Ada-002 embeddings with cosine similarity. \retmod{} runs up to $k=3$ iterations with base retrieval size $k_{\text{top}}=10$, confidence thresholds $\theta=0.7$ (general) and $\theta=0.85$ (temporal), and entity fact threshold $\delta=2$. Temperature is 0.3 for all LLM calls. All evaluations are zero-shot on the full LoCoMo benchmark.

\subsection{Main Results}

\subsubsection{Overall Performance}

Table~\ref{tab:results} presents results across all five categories.

\system{} achieves the highest overall scores and leads G-EVAL in four of five categories. MIRIX retains higher Judge Accuracy on single-hop, open-domain, and adversarial questions, reflecting stronger parametric knowledge, while \system{}'s largest gains appear in lexical overlap (F1, ROUGE-L, $+$56\% relative vs.\ MIRIX), confirming that iterative retrieval improves answer precision and recall.

\subsubsection{Category Analysis}

\paragraph{Temporal and multi-hop questions benefit most from evidence-gap-driven retrieval.}
Both categories require evidence distributed across sessions with no surface overlap---precisely the setting where single-pass retrieval fails and \retmod{}'s iterative gap-filling adds the most value.

\paragraph{Open-domain questions reveal a retrieval-first trade-off.}
MIRIX achieves higher Judge Accuracy on open-domain questions, despite \system{}'s superior F1.
MIRIX’s multi-agent design blends parametric knowledge with retrieved facts when evidence is sparse, while \retmod{} enforces strict grounding and abstains when conversational evidence is insufficient.

\paragraph{Gains scale with retrieval difficulty.}
Single-hop questions show modest gains, while multi-hop and temporal categories, requiring cross-session evidence, see the largest improvements. It confirms the benefits of evidence-gap–driven retrieval on harder tasks.

\begin{table*}[t]
\centering
\footnotesize
\setlength{\tabcolsep}{3pt}
\begin{tabular}{llccccc}
\toprule
\textbf{Category} & \textbf{Variant} & \textbf{G-EVAL} & \textbf{Judge Acc} & \textbf{F1} & \textbf{ROUGE-L} & \textbf{BERTScore} \\
\midrule
\textbf{Single-hop} & Single-pass & 2.51 & 61.3\% & 0.173 & 0.155 & 0.851 \\
& + Basic Loop & 2.63 & 61.8\% & 0.121 & 0.110 & 0.836 \\
& + Tiered Sufficiency & 2.73 & 65.7\% & 0.077 & 0.068 & 0.823 \\
& + Temporal Adapt. & 2.72 & 64.3\% & 0.081 & 0.072 & 0.824 \\
& \textbf{\system{} (Full)} & \textbf{2.98} & \textbf{68.2}\% & \textbf{0.205} & \textbf{0.187} & \textbf{0.857} \\
\midrule
\textbf{Multi-hop} & Single-pass & 2.67 & 81.4\% & 0.151 & 0.138 & 0.838 \\
& + Basic Loop & 2.68 & 80.1\% & 0.136 & 0.124 & 0.835 \\
& + Tiered Sufficiency & 3.09 & 87.3\% & 0.107 & 0.096 & 0.825 \\
& + Temporal Adapt. & 2.65 & 79.4\% & 0.281 & 0.255 & 0.850 \\
& \textbf{\system{} (Full)} & \textbf{2.89} & \textbf{85.2}\% & \textbf{0.260} & \textbf{0.239} & \textbf{0.857} \\
\midrule
\textbf{Temporal} & Single-pass & 1.87 & 58.8\% & 0.069 & 0.061 & 0.839 \\
& + Basic Loop & 3.09 & 82.7\% & 0.061 & 0.055 & 0.825 \\
& + Tiered Sufficiency & 3.03 & 83.7\% & 0.067 & 0.059 & 0.821 \\
& + Temporal Adapt. & 3.00 & 79.6\% & 0.070 & 0.063 & 0.822 \\
& \textbf{\system{} (Full)} & \textbf{3.08} & \textbf{81.6}\% & \textbf{0.135} & \textbf{0.124} & \textbf{0.845} \\
\midrule
\textbf{Open-domain} & Single-pass & 2.60 & 75.5\% & 0.164 & 0.162 & 0.852 \\
& + Basic Loop & 2.76 & 76.4\% & 0.129 & 0.128 & 0.841 \\
& + Tiered Sufficiency & 3.02 & 82.0\% & 0.114 & 0.115 & 0.835 \\
& + Temporal Adapt. & 2.90 & 79.4\% & 0.122 & 0.124 & 0.836 \\
& \textbf{\system{} (Full)} & \textbf{3.17} & \textbf{85.9}\% & \textbf{0.192} & \textbf{0.191} & \textbf{0.857} \\
\midrule
\textbf{Adversarial} & Single-pass & 1.17 & 43.4\% & 0.023 & 0.024 & 0.830 \\
& + Basic Loop & 1.48 & 48.3\% & 0.047 & 0.051 & 0.826 \\
& + Tiered Sufficiency & 1.86 & 51.7\% & 0.038 & 0.040 & 0.818 \\
& + Temporal Adapt. & 1.84 & 50.3\% & 0.041 & 0.043 & 0.819 \\
& \textbf{\system{} (Full)} & \textbf{1.94} & \textbf{55.1}\% & \textbf{0.081} & \textbf{0.083} & \textbf{0.840} \\
\bottomrule
\end{tabular}
\caption{\retmod{} additive ablation by question category. Each row adds one component group to the row above. All variants use the full \memmod{} memory.}
\label{tab:ablation_category}
\end{table*}

\subsubsection{Architectural Comparison with MIRIX}

MIRIX routes queries to specialized memory modules via LLM tool-calling, relying on the model to autonomously detect information gaps—a meta-cognitive task current models handle inconsistently. When initial retrieval misses key facts, it often proceeds with incomplete information.
\retmod{} instead makes gap detection \emph{explicit}: sufficiency evaluation diagnoses missing evidence and guides targeted re-retrieval.
This loop uses 2–6 LLM calls per question (each with a distinct role), compared to MIRIX’s variable tool-calling chains, yielding 4.5$\times$ lower latency (9.54s vs.\ 42.71s).

\subsection{Ablation Study}
\label{sec:ablation}

We conduct two additive ablation studies to isolate the contributions of the iterative retrieval framework (\retmod{}) and the layered memory structure (\memmod{}).
The first progressively adds \retmod{} components on top of single-pass retrieval over the full \memmod{} architecture (Table~\ref{tab:ablation_category}).
The second varies \memmod{}’s layered structure while holding \retmod{} fixed at its full configuration (Table~\ref{tab:lacemem_ablation}).

\subsubsection{System Variants}

\paragraph{\retmod{} variants.}
We define five variants in order of increasing capability:
\begin{enumerate}[leftmargin=*,nosep]
    \item \textbf{Single-pass} (\memmod{} only): one-pass retrieval with graph expansion and direct answer generation; no iteration.
    \item \textbf{+ Basic Loop}: adds iterative retrieval with single-level sufficiency evaluation and query refinement.
    \item \textbf{+ Tiered Sufficiency}: introduces three-tier evidence classification (\textsc{exact} / \textsc{inferrable} / \textsc{partial}) with approximate reasoning.
    \item \textbf{+ Temporal Adaptation}: adds temporal question detection \& adaptive confidence thresholds.
    \item \textbf{\system{} (Full)}: further adds entity tracking, anchor/refinement dual-path retrieval, and ``not mentioned’’ detection.
\end{enumerate}

\paragraph{\memmod{} variants.}
All use the full \retmod{} iterative loop; only memory structure changes:
\begin{enumerate}[leftmargin=*,nosep]
    \item \textit{Raw Only}: retrieves verbatim dialogue records via BM25; no structured indexing or graph expansion.
\item \textit{+ Index Layer}: adds the semantic Index layer (entity–predicate–object tuples with embedding retrieval) but disables Edge-layer expansion.
\item \textit{\memmod{} (Full)}: adds the Edge layer for relational graph traversal, enabling multi-hop expansion across memory nodes.
\end{enumerate}

\subsubsection{Results and Analysis}

Tables~\ref{tab:ablation_category} and~\ref{tab:lacemem_ablation} show category-wise results for the \retmod{} and \memmod{} ablations, respectively.

\begin{table*}[t]
\centering
\footnotesize
\setlength{\tabcolsep}{4pt}
\begin{tabular}{llccccc}
\toprule
\textbf{Category} & \textbf{Variant} & \textbf{G-EVAL} & \textbf{Judge Acc} & \textbf{F1} & \textbf{ROUGE-L} & \textbf{BERTScore} \\
\midrule
\textbf{Single-hop} & Raw Only & 2.69 & 20.3\% & 0.172 & 0.162 & 0.853 \\
& + Index Layer & 2.86 & 24.6\% & 0.197 & 0.183 & 0.855 \\
& \textbf{\memmod{} (Full)} & \textbf{2.98} & \textbf{68.2}\% & \textbf{0.205} & \textbf{0.187} & \textbf{0.857} \\
\midrule
\textbf{Multi-hop} & Raw Only & 1.82 & 16.7\% & 0.108 & 0.107 & 0.838 \\
& + Index Layer & 2.89 & 43.2\% & 0.273 & 0.250 & 0.860 \\
& \textbf{\memmod{} (Full)} & \textbf{2.89} & \textbf{85.2}\% & \textbf{0.260} & \textbf{0.239} & \textbf{0.857} \\
\midrule
\textbf{Temporal} & Raw Only & 2.98 & 45.8\% & 0.135 & 0.129 & 0.847 \\
& + Index Layer & 3.07 & 44.8\% & 0.137 & 0.126 & 0.848 \\
& \textbf{\memmod{} (Full)} & \textbf{3.08} & \textbf{81.6}\% & \textbf{0.135} & \textbf{0.124} & \textbf{0.845} \\
\midrule
\textbf{Open-domain} & Raw Only & 3.58 & 57.4\% & 0.239 & 0.239 & 0.864 \\
& + Index Layer & 3.14 & 42.3\% & 0.192 & 0.193 & 0.858 \\
& \textbf{\memmod{} (Full)} & \textbf{3.17} & \textbf{85.9}\% & \textbf{0.192} & \textbf{0.191} & \textbf{0.857} \\
\midrule
\textbf{Adversarial} & Raw Only & 2.40 & 29.8\% & 0.140 & 0.144 & 0.849 \\
& + Index Layer & 1.91 & 14.8\% & 0.084 & 0.086 & 0.841 \\
& \textbf{\memmod{} (Full)} & \textbf{1.94} & \textbf{55.1}\% & \textbf{0.081} & \textbf{0.083} & \textbf{0.840} \\
\bottomrule
\end{tabular}
\caption{\memmod{} ablation by question category. All variants use the full \retmod{} iterative loop; only the memory structure varies.}
\label{tab:lacemem_ablation}
\end{table*}

\paragraph{Effect of iterative retrieval (\retmod{}).}
The basic iterative loop yields the largest single-step gain on temporal questions (Judge Accuracy: 58.8\%\,$\to$\,82.7\%), confirming that closed-loop retrieval is essential when evidence is distributed across sessions.
Tiered sufficiency further improves multi-hop accuracy (80.1\%\,$\to$\,87.3\%) by preventing premature termination via the \textsc{inferrable} tier.
Temporal adaptation improves lexical precision on time-sensitive questions (multi-hop F1: 0.107\,$\to$\,0.281) at the cost of conservatism elsewhere.
The final component group---entity tracking, dual-path retrieval, and ``not mentioned’’ detection---contributes the largest single-hop F1 gain (0.081\,$\to$\,0.205) and raises adversarial accuracy from 50.3\% to 55.1\%.

\paragraph{Effect of memory structure (\memmod{}).}
The Index layer improves multi-hop Judge Accuracy from 16.7\% to 43.2\% via atomic tuple matching, but without the Edge layer introduces fragmentation (Open-domain: 57.4\%\,$\to$\,42.3\%; Adversarial: 29.8\%\,$\to$\,14.8\%).
Adding the Edge layer resolves this: multi-hop accuracy reaches 85.2\%, temporal 81.6\%, and adversarial recovers to 55.1\%, confirming that graph-based expansion is necessary for assembling coherent evidence sets.

\paragraph{Joint effects.} The two ablations show that \retmod{} and \memmod{} are mutually reinforcing: iterative gap-filling requires fine-grained structured evidence, while the layered memory realises its potential only under evidence-gap-driven retrieval.

\paragraph{Diagnosis-driven vs.\ generic re-query.}
To isolate the contribution of the diagnosis signal from iteration itself, we compare \retmod{}'s refinement against a generic re-query baseline that generates a new query at each iteration without access to the gap diagnosis. At iteration~2, on questions unresolved after iteration~1, diagnosis-driven refinement improves Recall@5 by +2.6\%, Recall@10 by +2.0\%, and nDCG@10 by +2.2\% over generic re-query, with all six retrieval metrics consistently positive (full per-iteration breakdown in Appendix Table~\ref{tab:retrieval_per_iter}). By iteration~3 the recall gap narrows but nDCG remains higher (+0.5\%), indicating that the diagnosis signal continues to improve ranking quality. While these retrieval-level gains are modest in isolation, the sufficiency evaluator amplifies them into the substantially larger end-to-end improvements reported in Tables~\ref{tab:results}--\ref{tab:ablation_category}: a small increase in gold-evidence recall translates, through more accurate tier classification and earlier termination, into +15.2\% Judge Accuracy over single-pass retrieval.

\paragraph{Mechanisms behind non-monotone ablation trajectories.}
Adding tiered sufficiency causes Multi-hop F1 to fall from 0.151 to 0.107 while Judge Accuracy rises from 80.1\% to 87.3\%: the \textsc{inferrable} tier produces semantically correct but inference-bridged answers (``Based on facts A and B, the answer is C'') that diverge from gold annotations' concise phrasing. The subsequent jump to 0.281 under Temporal Adaptation reflects LoCoMo's multi-hop question distribution: many such questions involve temporal reasoning, and explicit calendar anchoring aligns answers with gold formatting. The drop-then-recovery pattern holds across all ten conversations (10/10).

The Open-domain drop under Index-only (57.4\%\,$\to$\,42.3\%) reflects loss of relational context: raw-text retrieval returns conversational passages where preferences are naturally co-located, while atomic tuples without Edge connections produce isolated facts that cannot support synthesis questions like ``What kind of person is X?''. The fragmentation does not harm Single-hop or Multi-hop, where precise fact targeting dominates, but degrades Open-domain by removing aggregate descriptions. The Edge layer restores performance (85.9\%) by reconnecting same-source tuples; the pattern holds across all ten conversations (10/10 drop without Edge).

\FloatBarrier
\begin{table*}[t]
\centering
\begin{minipage}[t]{0.42\textwidth}
\centering
\footnotesize
\setlength{\tabcolsep}{3pt}
\begin{tabular}{lcccc}
\toprule
 & Binary & Prec. & Recall & EXACT \\
\midrule
Overall      & 65.0 & 89.0 & 62.4 & 94.6 \\
\midrule
Single-hop   & 79.4 & 91.2 & 82.3 & 76.9 \\
Multi-hop    & 48.7 & 95.5 & 44.8 & 98.8 \\
Temporal     & 65.4 & 96.2 & 63.3 & 100.0 \\
Open-domain  & 74.3 & 92.3 & 76.0 & 98.8 \\
Adversarial  & 57.1 & 72.2 & 42.8 & 37.5 \\
\bottomrule
\end{tabular}
\caption{Sufficiency classifier agreement with the oracle on the full LoCoMo benchmark (\%). Binary Agr.\ collapses tiers into sufficient vs.\ insufficient; EXACT Prec.\ is the precision of \textsc{exact} commitments.}
\label{tab:sufficiency_validation}
\end{minipage}
\hfill
\begin{minipage}[t]{0.55\textwidth}
\centering
\footnotesize
\resizebox{\linewidth}{!}{%
\begin{tabular}{lrrrrrr}
\toprule
\textbf{System / Variant} & \textbf{Avg} & \textbf{Single} & \textbf{Multi} & \textbf{Temp.} & \textbf{Open} & \textbf{Adv.} \\
\midrule
MIRIX & 42.71 & 38.59 & 42.64 & 43.79 & 40.71 & 48.92 \\
\textbf{\system{} (Full)} & \textbf{9.54} & \textbf{8.56} & \textbf{12.11} & \textbf{9.07} & \textbf{10.72} & \textbf{8.63} \\
\midrule
\multicolumn{7}{l}{\textit{\retmod{} ablation (full \memmod{})}} \\
\midrule
\quad Single-pass & 2.13 & 2.21 & 1.99 & 2.21 & 2.59 & 2.10 \\
\quad + Basic Loop & 5.73 & 5.14 & 7.34 & 5.04 & 6.99 & 5.18 \\
\quad + Tiered Suff. & 8.04 & 7.50 & 10.24 & 7.68 & 9.72 & 7.02 \\
\quad + Temporal Adapt. & 7.84 & 7.35 & 9.37 & 8.20 & 8.77 & 6.95 \\
\midrule
\multicolumn{7}{l}{\textit{\memmod{} ablation (full \retmod{})}} \\
\midrule
\quad Raw Only & 4.83 & 4.79 & 5.06 & 5.58 & 4.35 & 5.44 \\
\quad + Index Layer & 6.58 & 5.89 & 6.10 & 6.64 & 6.06 & 8.34 \\
\bottomrule
\end{tabular}}
\caption{Inference latency (s/question) overall and by question category. The category variants are Single-hop, Multi-hop, Temporal, Open-domain, Adversarial.}
\label{tab:efficiency_combined}
\end{minipage}
\end{table*}

\paragraph{Independent judge validation.}
To address potential self-preference bias from using GPT-4o as both answer generator and judge, we re-evaluate all \system{} outputs on the full LoCoMo benchmark using DeepSeek-V3.2. The two judges agree on 90.6\% of decisions and their accuracy estimates differ by only 1.3~pp (GPT-4o 76.5\% vs.\ DeepSeek 75.2\%). The sign of the gap varies across categories (DeepSeek is more lenient on three of five categories; see Appendix Table~\ref{tab:robustness_per_category}), indicating no systematic self-preference bias. The largest disagreement occurs on Multi-hop (agreement 83.3\%), where DeepSeek is more strict about synthesis-style answers; since all baselines share the same judge, the relative ranking between \system{} and baselines is unaffected.

\paragraph{LLM backbone and embedding robustness.}
To evaluate \system{}'s sensitivity to LLM backbone and embedding choice, we run two independent substitution experiments on the full LoCoMo benchmark: \textit{(i)} replace all five internal LLM calls (answer generation, sufficiency evaluation, query refinement, entity extraction, and reasoning chain) with DeepSeek-V3.2; \textit{(ii)} replace the \texttt{text-embedding-3-small} index with \texttt{BAAI/bge-m3}, an open-weights multilingual embedding model.

Each configuration is judged twice (GPT-4o-mini and DeepSeek-V3.2) to control for evaluator-side bias. Under the GPT-4o judge, the LLM-backbone swap costs 1.2~pp (76.5\%\,$\to$\,75.3\%) and the embedding swap 1.0~pp (76.5\%\,$\to$\,75.5\%); under the DeepSeek judge, 3.9~pp and 2.1~pp respectively (Appendix Table~\ref{tab:robustness_matrix}). The two judges agree on the qualitative pattern: backbone substitution is slightly costlier than embedding substitution, and cross-family agreement stays above 90\% in every cell (90.6\% / 92.5\% / 91.5\%). The per-category profile is preserved (each category shifts by $\leq 4$~pp; Appendix Table~\ref{tab:robustness_per_category}), and bge-m3 actually improves Multi-hop accuracy by +3.8~pp over the OpenAI embedding. \system{}'s improvements therefore depend on neither a specific LLM family nor a specific embedding model.

\paragraph{Iterative graph expansion.}
\retmod{} performs multi-hop reasoning through iteration, not through fixed-depth graph traversal. Each round's sufficiency diagnosis identifies a missing evidence direction; the refined query selects fresh seed nodes; Edge expansion adds their immediate neighbours to the cumulative pool. Across three rounds, this reaches deeper into the graph than any single fixed-radius traversal while keeping every step semantically tight: each hop follows a direction the sufficiency signal has already validated. Static multi-hop expansion, in contrast, lacks this per-hop diagnostic guidance. Table~\ref{tab:lacemem_ablation} confirms Edge expansion's importance: enabling it raises Multi-hop Judge Accuracy from 43.2\% to 85.2\% and Adversarial accuracy from 14.8\% to 55.1\%.

\subsection{Sufficiency Classifier Validation}
\label{sec:sufficiency_validation}

To verify that \retmod{}'s tiered sufficiency signal is reliable independently of downstream answer quality, we construct an oracle from LoCoMo's annotated \texttt{evidence} field (the \texttt{dia\_id}s containing each answer). An iteration is labelled \textsc{exact} if every annotated \texttt{dia\_id} is covered, \textsc{inferrable} if the retrieved facts can still support inference of the gold answer, and \textsc{partial} or \textsc{none} otherwise. Grounding the oracle in retrieval coverage decouples the classifier's judgement from the generator: the same retrieved evidence yields the same oracle label.

When the classifier declares evidence sufficient, it agrees with the oracle in 89.0\% of cases. When it commits to the strictest \textsc{exact} tier (the only signal that terminates iteration in \retmod{}), its precision against the oracle reaches 94.6\%, rising to 95.5\% on Multi-hop questions where evidence is spread across sessions. The classifier is therefore reliable in its broad sufficiency judgement and near-perfect when it commits to its strongest termination signal.

\paragraph{Threshold-level commitment reliability.}
\retmod{}'s termination rule is threshold-based, so we evaluate the classifier at the decision boundaries it actually uses. Confidence is highly monotonic with oracle sufficiency (Spearman $\rho = 0.93$). At the \textsc{exact} termination threshold, commitments reach Precision@0.85 = 97.9\%; at the \textsc{inferrable} threshold, Precision@0.7 = 91.3\%. PR-AUC is 0.907 (vs.\ 0.774 random). \retmod{} therefore terminates only on evidence it has correctly identified as sufficient.

\paragraph{Failure analysis: adversarial and open-domain categories.}
We classify wrong answers on Cat\,5 (Adversarial, $N$=446) and Cat\,4 (Open-domain, $N$=841) into three types: \emph{commission errors} (wrong non-abstention answer), \emph{false abstentions} (``not mentioned'' when an answer exists), and \emph{missed abstentions} (answers when it should abstain). Commission errors dominate both categories, with the dominance more pronounced on Cat\,4; false abstentions form a smaller share (Cat\,4 < Cat\,5), and missed abstentions are negligible.

\retmod{} rarely commits errors at high confidence: only 0.5\% (Cat\,4) and 1.3\% (Cat\,5) of commissions occur at \textsc{exact} or with confidence $\geq 0.85$, consistent with the 94.6\% \textsc{exact} precision in \S\ref{sec:sufficiency_validation}. False abstentions are equally constrained: 85.2\% (Cat\,4) and 92.6\% (Cat\,5) occur only after the three-iteration budget is exhausted, and all sit at the \textsc{partial} tier, identifying abstention as the loop's terminal fallback.

Adversarial questions in LoCoMo almost always have a real answer (only 2 of 446 are gold ``Not~mentioned''); the challenge is retrieval \emph{precision}, not absence detection. Many Cat\,4 near-miss cases (ROUGE-L\,$>$\,0.4, judge\,=\,0) reflect partial retrieval: \retmod{} assembles some of the required evidence but misses a completing detail, producing a plausible but incomplete answer.

\subsection{Efficiency Analysis}

Table~\ref{tab:efficiency_combined} reports latency per question overall and by category. \system{} is 4.5$\times$ faster than MIRIX (9.54s vs.\ 42.71s) while maintaining comparable or higher accuracy. Each \retmod{} component adds moderate overhead over Single-pass (2.13s), with incremental cost from the iterative loop. Latency scales with question complexity: single-hop questions converge quickly (8.56s), while multi-hop require more iterations (12.11s), allocating computation proportional to difficulty.

\section{Conclusion}
We present \system{}, including \memmod{}’s coarse-to-fine memory hierarchy with \retmod{}’s evidence-gap–driven iterative retrieval.
\memmod{} structures dialogue history into Index, Edge, and Raw layers for efficient search, expansion, and grounding.
\retmod{} iteratively retrieves by evaluating evidence sufficiency, diagnosing gaps, and refining queries.
On LoCoMo, \system{} achieves 81.6\% Judge Accuracy on temporal questions (vs.\ 73.3\% MIRIX, 58.8\% single-pass) and 85.2\% on multi-hop (vs.\ 65.9\% MIRIX), while running 4.5$\times$ faster than MIRIX. Additive ablations show all components contribute incrementally. The key insight is that making evidence insufficiency explicit enables targeted gap-filling missed by single-pass and implicit-feedback methods.

\raggedbottom

\section*{Limitations}
While \system{} demonstrates strong performance, one limitation warrants future exploration. \memmod{}'s layered memory is currently constructed from a complete conversation snapshot, following the standard offline evaluation protocol of LoCoMo. Although the architecture's modular design---with separate Index, Edge, and Raw layers---is naturally amenable to incremental updates, extending memory construction to an online, streaming setting where new dialogue turns are ingested and indexed in real time remains an important direction for practical deployment.

\bibliography{custom}

\appendix

\section{Detailed \retmod{} Algorithm}
\label{app:algo}

Algorithm~\ref{alg:irr_loop_detailed} provides the full specification of the \retmod{} iterative retrieval loop summarised in Algorithm~\ref{alg:irr_loop}, including all confidence constants, entity-aware adjustments, termination thresholds, and the query-refinement prompt template.

\begin{algorithm*}[t]
\small
\SetAlgoVlined
\SetAlgoSkip{}
\caption{\retmod{} --- Detailed Iterative Retrieval Loop with Confidence Constants and Thresholds}
\label{alg:irr_loop_detailed}
\KwIn{Question $q$, Memory $\mathcal{M}$, Entity set $\mathcal{E}$, Max iterations $k$, Entity fact threshold $\delta$}
\KwOut{Evidence $\mathcal{D}_{all}$, Confidence $c$, Evidence tier $t$}
$q' \gets q$;\; $\mathcal{D}_{all} \gets \emptyset$;\; $\mathcal{F}_e \gets \emptyset\;\forall\, e \in \mathcal{E}$\;

\For{$i \gets 1$ \KwTo $k$}{
    \tcp{\textbf{Phase 1: Dual-Path Retrieval} --- anchor preserves coverage, refinement targets gaps}
    $k_{ret} \gets 10+3(i-1)$ \tcp*{Progressive widening of retrieval scope}
    $\mathcal{D}_{anc} \gets \texttt{Retrieve}(q,\; k_{ret})$ \tcp*{Anchor path: original question}
    $\mathcal{D}_{ref} \gets \texttt{Retrieve}(q',\; k_{ret})$ \tcp*{Refinement path: diagnosis-driven query}
    $\mathcal{D}^{(i)} \gets \texttt{Dedup}\!\big(\mathcal{D}_{anc} \cup \mathcal{D}_{ref} \cup \texttt{GraphExpand}(\mathcal{D}_{anc} \cup \mathcal{D}_{ref})\big)$ \tcp*{Expand via \memmod{} Edge layer}
    $\mathcal{D}_{all} \gets \mathcal{D}_{all} \cup \mathcal{D}^{(i)}$ \tcp*{Accumulate evidence across iterations}

    \BlankLine
    \tcp{\textbf{Phase 2: Entity Tracking} --- monitor per-entity fact coverage}
    \ForEach{entity $e \in \mathcal{E}$}{
        $\mathcal{F}_e \gets \mathcal{F}_e \cup \{d \in \mathcal{D}^{(i)} \mid \texttt{Contains}(d, e)\}$\;
    }

    \BlankLine
    \tcp{\textbf{Phase 3: Sufficiency Evaluation} --- assess accumulated evidence as a whole}
    $(t,\, c,\, m) \gets \texttt{EvalSufficiency}(\mathcal{D}_{all},\, q)$ \tcp*{$t \in \{\textsc{exact}, \textsc{inferrable}, \textsc{partial}\}$; $m$ = missing info}
    \If(\tcp*[f]{Temporal questions require stricter thresholds}){$\tau(q) = \texttt{True}$}{
        $c \gets \begin{cases} \max(c, 0.85) & \text{if } t = \textsc{exact} \\ \min(c, 0.75) & \text{if } t = \textsc{inferrable} \\ \min(c, 0.50) & \text{if } t = \textsc{partial} \end{cases}$\;
    }
    \If(\tcp*[f]{Downgrade if any entity lacks sufficient evidence}){$\exists\, e \in \mathcal{E}:\; |\mathcal{F}_e| < \delta$}{
        $t \gets \textsc{partial}$\;
        $c \gets \min(c, 0.6)$\;
        $m \gets m \,\oplus$ ``need more about: $\{e \mid |\mathcal{F}_e| < \delta\}$''\;
    }

    \BlankLine
    \tcp{\textbf{Phase 4: Termination} --- stop when sufficient or budget exhausted}
    \lIf{$\texttt{Sufficient}(t,\, c)$ \textbf{or} $i = k$}{\textbf{break}}

    \BlankLine
    \tcp{\textbf{Phase 5: Diagnosis-Driven Query Refinement} --- use $m$ to generate next query}
    \tcp{Step 5a: Strategy selection (rule-based, no LLM call)}
    \eIf{$\tau(q) = \texttt{True}$}{
        $s \gets \texttt{TemporalStrategy}(i)$ \tcp*{Temporal-specific refinement strategy}
    }{
        $s \gets \texttt{GeneralStrategy}(i)$ \tcp*{e.g., keywords $\to$ different angle $\to$ synonyms}
    }
    \tcp{Step 5b: Entity context injection (only for under-represented entities)}
    $\varepsilon \gets \texttt{``''}$\;
    \If{$\exists\, e \in \mathcal{E}:\; |\mathcal{F}_e| < \delta$}{
        $\varepsilon \gets$ ``Need more about: $\{e\,(|\mathcal{F}_e|) \mid |\mathcal{F}_e| < \delta\}$. Include entity names in query.''\;
    }
    \tcp{Step 5c: Prompt assembly and LLM query generation}
    $\textit{prompt} \gets$ \fbox{\parbox{0.55\linewidth}{\ttfamily\footnotesize
        Original question: $q$ \\
        Current search query: $q'$ \\
        Missing information: $m$ \\
        Iteration: $i$\,/\,$k$ \\[2pt]
        Strategy: $s$ \\
        Entity context: $\varepsilon$ \hfill {\normalfont\itshape (only if $\varepsilon \neq$ ``'')} \\[2pt]
        Generate an improved search query. Return ONLY the query.
    }}\;
    $q' \gets \texttt{LLM}(\textit{prompt})$\;
}
\Return{$\mathcal{D}_{all},\; c,\; t$}\;
\end{algorithm*}

\section{Case Study}
\label{app:case_study}

We present four pipeline traces showing each step of \retmod{}.

\begin{table*}[t]
\centering
\small
\begin{tabularx}{\textwidth}{p{2.5cm} X}
\toprule
\textbf{Question} & How do Jon and Gina both like to destress? (Cat.\ 4, General) \\
\textbf{Ground Truth} & by dancing \\
\midrule
\textbf{Entity Extraction} & Regex match $\rightarrow$ Entities: [Jon, Gina]; Type: GENERAL \\
\midrule
\textbf{Iter 1: Retrieval} & Query: ``How do Jon and Gina both like to destress?'' \newline Index $\rightarrow$ 10 facts; Edge $\rightarrow$ +15 facts \newline Top facts: \newline \quad \texttt{Gina advises take breaks and dance to destress} \newline \quad \texttt{Jon is collaborating with Gina} \newline \quad \texttt{Gina supports Jon} \newline Per-entity: Jon $\rightarrow$ 8 facts; Gina $\rightarrow$ 12 facts \\
\textbf{Iter 1: Sufficiency} & \textit{Prompt $\rightarrow$ LLM:} \newline \texttt{Question: How do Jon and Gina both like to destress?} \newline \texttt{Retrieved Facts: [25 facts total]} \newline \texttt{Evaluate: EXACT/INFERRABLE/PARTIAL/CONFIDENCE/MISSING} \newline \textit{LLM Response:} \newline \texttt{PARTIAL: yes, CONFIDENCE: 0.4} \newline \texttt{MISSING: how Jon destresses; only Gina's method found} \newline Entity check: Jon has 8 facts $<$ threshold $\rightarrow$ downgrade, $c=0.4$ \\
\textbf{Iter 1: Refinement} & \textit{Prompt $\rightarrow$ LLM:} \newline \texttt{Original: How do Jon and Gina both like to destress?} \newline \texttt{Missing: how Jon destresses} \newline \texttt{Strategy: Focus on keywords, entity names, key concepts.} \newline \texttt{IMPORTANT: Need more about: Jon (only 8 facts)} \newline \textit{LLM Response:} ``Jon stress relief relaxation hobby'' \\
\midrule
\textbf{Iter 2: Retrieval} & Refined query $\rightarrow$ 26 new facts via Edge expansion \newline Key: \texttt{Dancing helps Jon de-stress}, \texttt{Jon finds stress relief in dancing} (Edge), \texttt{Gina uses\_for\_stress\_relief dance} (Edge) \newline Per-entity: Jon $\rightarrow$ 40 facts; Gina $\rightarrow$ 25 facts \\
\textbf{Iter 2: Sufficiency} & \textit{LLM Response:} \newline \texttt{EXACT: no} \newline \texttt{INFERRABLE: yes} \newline \texttt{CONFIDENCE: 0.8} \newline \texttt{MISSING: none} \newline $\Rightarrow$ \textsc{inferrable}, $c=0.8 \geq 0.7$ $\rightarrow$ \textit{Terminate} \\
\midrule
\textbf{Reasoning Chain} & \textit{Prompt $\rightarrow$ LLM:} \newline \texttt{Analyze if multi-hop reasoning is needed.} \newline \texttt{Question: How do Jon and Gina both like to destress?} \newline \texttt{Entities: Gina, Jon; Facts: [15 facts]} \newline \textit{LLM Response:} \texttt{TYPE: MULTI-HOP} \newline \texttt{Step 1: Identify how Gina destresses} \newline \texttt{Step 2: Identify how Jon destresses} \newline \texttt{Step 3: Connect their methods} \\
\midrule
\textbf{Answer Generation} & \textit{Prompt $\rightarrow$ LLM:} \newline \texttt{Answer directly and confidently. Do NOT use hedging phrases.} \newline \texttt{Reasoning steps: 1. Gina's method 2. Jon's method 3. Connect} \newline \texttt{Question: How do Jon and Gina both like to destress?} \newline \texttt{Relevant Facts: [49 unique facts]} \newline \textit{LLM Response:} \newline \textbf{Answer:} \underline{Jon and Gina both like to destress by dancing.} \quad $c=0.8$ \quad \textcolor{darkgreen}{Correct} \\
\bottomrule
\end{tabularx}
\caption{Case B: Multi-Hop Entity Tracking (2 Iterations). Iteration 1 finds Gina's method but insufficient facts for Jon. Entity-aware refinement targets Jon, and iteration 2 discovers the shared stress-relief method via Edge expansion.}
\label{tab:case_b}
\end{table*}

\begin{table*}[t]
\centering
\small
\begin{tabularx}{\textwidth}{p{2.5cm} X}
\toprule
\textbf{Question} & When is Jon's group performing at a festival? (Cat.\ 2, Temporal) \\
\textbf{Ground Truth} & February, 2023 \\
\midrule
\textbf{Entity Extraction} & Regex match $\rightarrow$ Entities: [Jon]; Type: TEMPORAL \\
\midrule
\textbf{Iter 1: Retrieval} & Query: ``When is Jon's group performing at a festival?'' \newline Index $\rightarrow$ 8 facts; Edge $\rightarrow$ 41 facts \newline Key: \texttt{Jon performed\_at festival} (no date) \\
\textbf{Iter 1: Sufficiency} & \textit{Prompt $\rightarrow$ LLM:} \newline \texttt{Question: When is Jon's group performing at a festival?} \newline \texttt{Retrieved Facts:} \newline \texttt{- Jon performed\_at festival} \newline \texttt{- Jon loves performing} \newline \texttt{- Jon is rehearsing for upcoming show} \newline \texttt{...\ [20 facts total]} \newline \texttt{IMPORTANT: This is a TEMPORAL question asking for specific dates/times.} \newline \texttt{- EXACT\_MATCH: Has precise date/time} \newline \texttt{- INFERRABLE: Has temporal clues for inference} \newline \texttt{- PARTIAL\_MATCH: Related but insufficient} \newline \texttt{Evaluate: EXACT/INFERRABLE/PARTIAL/CONFIDENCE/MISSING} \newline \textit{LLM Response:} \newline \texttt{EXACT: no} \newline \texttt{INFERRABLE: no} \newline \texttt{PARTIAL: yes} \newline \texttt{CONFIDENCE: 0.3} \newline \texttt{MISSING: Specific date or time of Jon's performance at the festival} \\
\textbf{Iter 1: Refinement} & \textit{Prompt $\rightarrow$ LLM:} \newline \texttt{Original question: When is Jon's group performing at a festival?} \newline \texttt{Current search query: When is Jon's group performing at a festival?} \newline \texttt{Missing information: specific date or time of Jon's performance} \newline \texttt{Iteration: 1/3} \newline \texttt{Strategy: Focus on DATE, TIME, and temporal keywords (when, started, launched, opened).} \newline \texttt{Generate an improved search query. Return ONLY the query.} \newline \textit{LLM Response:} \newline ``Jon group festival performance date February'' \\
\midrule
\textbf{Iter 2: Retrieval} & Refined query retrieves 6 new facts via Edge expansion \newline Key: \texttt{Jon group is\_performing\_at festival (time: 2023-02)} \\
\textbf{Iter 2: Sufficiency} & \textit{LLM Response:} \newline \texttt{EXACT: no} \newline \texttt{INFERRABLE: yes} \newline \texttt{CONFIDENCE: 0.75} \newline \texttt{MISSING: none} \newline $\Rightarrow$ \textsc{inferrable}, $c=0.75$ (capped for temporal) $\rightarrow$ \textit{Terminate} \\
\midrule
\textbf{Answer Generation} & \textit{Prompt $\rightarrow$ LLM:} \newline \texttt{The facts contain temporal clues that allow REASONABLE INFERENCE.} \newline \texttt{Make a careful inference:} \newline \texttt{- If ``as of February 2023'', the event likely happened in/around February 2023} \newline \texttt{- Be clear but concise. Use ``in'' or ``around'' if inferring timeframe} \newline \texttt{- Do NOT be overly cautious with phrases like ``it is likely that''} \newline \texttt{Question: When is Jon's group performing at a festival?} \newline \texttt{Relevant Facts:} \newline \texttt{- Jon group is\_performing\_at festival (time: 2023-02)} \newline \texttt{- Jon performed\_at festival} \newline \texttt{...\ [46 unique facts]} \newline \texttt{Answer (be concise and direct):} \newline \textit{LLM Response:} \newline \textbf{Answer:} \underline{February 2023} \quad $c=0.75$ \quad \textcolor{darkgreen}{Correct} \\
\bottomrule
\end{tabularx}
\caption{Case C: Temporal Inference (2 Iterations). Iteration 1 retrieves the event without a date. Query refinement targets temporal keywords, and iteration 2 finds the temporal tuple via Edge expansion.}
\label{tab:case_c}
\end{table*}

\begin{table*}[t]
\centering
\small
\begin{tabularx}{\textwidth}{p{2.5cm} X}
\toprule
\textbf{Question} & Why did Jon decide to start his dance studio? (Cat.\ 4, General) \\
\textbf{Ground Truth} & He lost his job and decided to start his own business to share his passion. \\
\midrule
\textbf{Entity Extraction} & Regex match $\rightarrow$ Entities: [Jon]; Type: GENERAL \\
\midrule
\textbf{Iter 1: Retrieval} & Query: ``Why did Jon decide to start his dance studio?'' \newline Index $\rightarrow$ 10 facts; Edge $\rightarrow$ +20 facts \newline Key facts: \newline \quad \texttt{Jon wants\_to\_start dance studio} \newline \quad \texttt{Jon is\_following\_passion\_for dance} \newline \quad \texttt{Jon is\_turning\_love\_of dance\_into\_business} \newline \quad \texttt{Jon wants studio to be a place of support} \\
\textbf{Iter 1: Sufficiency} & \textit{Prompt $\rightarrow$ LLM:} \newline \texttt{Question: Why did Jon decide to start his dance studio?} \newline \texttt{Retrieved Facts: [30 facts including passion-related tuples]} \newline \texttt{Evaluate: EXACT/INFERRABLE/PARTIAL/CONFIDENCE/MISSING} \newline \textit{LLM Response:} \newline \texttt{INFERRABLE: yes, CONFIDENCE: 0.55} \newline \texttt{MISSING: what triggered the decision beyond passion} \newline $\Rightarrow$ $c=0.55 < 0.7$ $\rightarrow$ continue \\
\textbf{Iter 1: Refinement} & \textit{Prompt $\rightarrow$ LLM:} \newline \texttt{Original: Why did Jon decide to start his dance studio?} \newline \texttt{Missing: what triggered the decision beyond passion} \newline \texttt{Strategy: Focus on keywords, entity names, key concepts.} \newline \textit{LLM Response:} ``Jon reason start dance studio job loss career change'' \\
\midrule
\textbf{Iter 2: Retrieval} & Refined query $\rightarrow$ +35 new facts via Edge expansion \newline Key: \texttt{Jon lost job} (Edge), \texttt{Jon transitioned from secure 9-5 job} (Edge) \\
\textbf{Iter 2: Sufficiency} & \textit{LLM Response:} \newline \texttt{EXACT: no} \newline \texttt{INFERRABLE: yes} \newline \texttt{CONFIDENCE: 0.8} \newline \texttt{MISSING: none} \newline $\Rightarrow$ \textsc{inferrable}, $c=0.8 \geq 0.7$ $\rightarrow$ \textit{Terminate} \\
\midrule
\textbf{Reasoning Chain} & \textit{Prompt $\rightarrow$ LLM:} \newline \texttt{Analyze if multi-hop reasoning is needed.} \newline \texttt{Question: Why did Jon decide to start his dance studio?} \newline \texttt{Entities: Jon; Facts: [15 facts]} \newline \textit{LLM Response:} \texttt{TYPE: MULTI-HOP} \newline \texttt{Step 1: Identify Jon's background and job loss} \newline \texttt{Step 2: Connect job loss to dance passion} \newline \texttt{Step 3: Identify studio motivations} \\
\midrule
\textbf{Answer Generation} & \textit{Prompt $\rightarrow$ LLM:} \newline \texttt{Answer directly and confidently. Do NOT use hedging phrases.} \newline \texttt{Reasoning steps: 1. Job loss 2. Dance passion 3. Studio motivations} \newline \texttt{Question: Why did Jon decide to start his dance studio?} \newline \texttt{Relevant Facts: [65 unique facts]} \newline \textit{LLM Response:} \newline \textbf{Answer:} ``Jon lost his job as a banker and decided to follow his passion for dancing by starting his own dance studio.'' \quad $c=0.8$ \quad \textcolor{darkgreen}{Correct} \\
\bottomrule
\end{tabularx}
\caption{Case D: Multi-Hop Reasoning Chain (2 Iterations). Iteration 1 finds passion-related facts but misses the job-loss trigger ($c=0.55$). Query refinement targets ``job loss career change'', and iteration 2 discovers the causal link via Edge expansion.}
\label{tab:case_d}
\end{table*}

\section{Per-Iteration Retrieval Metrics}
\label{app:retrieval_per_iter}

Table~\ref{tab:retrieval_per_iter} reports the full per-iteration retrieval metrics summarised in \S\ref{sec:ablation}, comparing diagnosis-driven refinement against the generic re-query baseline on the full LoCoMo benchmark.

\begin{table*}[t]
\centering
\footnotesize
\setlength{\tabcolsep}{4pt}
\begin{tabular}{llccccccc}
\toprule
\textbf{Variant} & \textbf{Iter} & \textbf{$N$} & \textbf{R@5} & \textbf{R@10} & \textbf{R@20} & \textbf{nDCG@5} & \textbf{nDCG@10} & \textbf{nDCG@20} \\
\midrule
\multirow{3}{*}{Generic re-query}
  & 1 & 1{,}982 & 50.0 & 56.8 & 59.7 & 42.4 & 44.8 & 45.7 \\
  & 2 &   578 & 28.1 & 33.8 & 38.3 & 23.9 & 25.8 & 27.0 \\
  & 3 &   486 & 29.3 & 34.0 & 38.0 & 24.6 & 26.1 & 27.2 \\
\midrule
\multirow{3}{*}{Diagnosis-driven (\retmod{})}
  & 1 & 1{,}982 & 50.0 & 56.8 & 59.7 & 42.4 & 44.8 & 45.7 \\
  & 2 &   578 & \textbf{30.7} & \textbf{35.8} & \textbf{39.8} & \textbf{26.3} & \textbf{28.0} & \textbf{29.2} \\
  & 3 &   486 & 29.1 & 33.8 & 38.1 & 25.0 & 26.6 & 27.8 \\
\bottomrule
\end{tabular}
\caption{Per-iteration retrieval metrics for diagnosis-driven refinement vs.\ generic re-query on the full LoCoMo benchmark. $N$ is the number of questions reaching that iteration; iter~$\geq$2 contains only those unresolved by earlier iterations. Iteration~1 numbers are identical because the diagnosis signal is unavailable until iteration~2. Bold cells highlight the iteration~2 row where diagnosis-driven retrieval beats generic re-query on all six metrics.}
\label{tab:retrieval_per_iter}
\end{table*}

\section{Robustness and Judge Validation}
\label{app:robustness}

Tables~\ref{tab:robustness_matrix} and~\ref{tab:robustness_per_category} report the robustness and judge-validation experiments referenced in \S\ref{sec:ablation}. Each row is judged twice (GPT-4o-mini and DeepSeek-V3.2): the GPT-4o columns serve the backbone-and-embedding substitution analysis, while the DeepSeek columns and cross-judge agreement support the independent judge validation.

\begin{table*}[t]
\centering
\footnotesize
\setlength{\tabcolsep}{4pt}
\begin{tabular}{lcccc}
\toprule
\textbf{Configuration} & \textbf{Judge Acc (GPT-4o)} & \textbf{Judge Acc (DeepSeek)} & \textbf{$\Delta$ vs.\ baseline} & \textbf{Cross-judge Agreement} \\
\midrule
OpenAI emb. + GPT-4o backbone (baseline) & 76.5\% & 75.2\% & ref. & 90.6\% \\
OpenAI emb. + DeepSeek backbone          & 75.3\% & 71.3\% & $-1.2$ / $-3.9$ pp & 92.5\% \\
bge-m3 emb. + GPT-4o backbone            & 75.5\% & 73.1\% & $-1.0$ / $-2.1$ pp & 91.5\% \\
\bottomrule
\end{tabular}
\caption{Robustness of \system{} to LLM backbone and embedding substitution. Each row reports overall Judge Accuracy on the full LoCoMo benchmark under both GPT-4o-mini and DeepSeek-V3.2 judges; the last column reports cross-judge agreement within each cell. Both substitution axes degrade overall accuracy by at most 1.2 pp under the GPT-4o judge.}
\label{tab:robustness_matrix}
\end{table*}

\begin{table*}[t]
\centering
\footnotesize
\setlength{\tabcolsep}{5pt}
\begin{tabular}{l|ccc|ccc}
\toprule
                     & \multicolumn{3}{c|}{\textbf{GPT-4o-mini judge}} & \multicolumn{3}{c}{\textbf{DeepSeek-V3.2 judge}} \\
\textbf{Category}    & \textbf{Baseline} & \textbf{+ DS bk} & \textbf{+ bge-m3} & \textbf{Baseline} & \textbf{+ DS bk} & \textbf{+ bge-m3} \\
\midrule
Single-hop  & 68.2 & 66.2 & 65.2 & 70.4 & 67.0 & 68.0 \\
Multi-hop   & 85.2 & 85.9 & \textbf{89.0} & 72.9 & 71.6 & 73.8 \\
Temporal    & 81.6 & 81.6 & 83.7 & 81.6 & 79.5 & 82.7 \\
Open-domain & 85.9 & 85.3 & 84.4 & 88.4 & 85.5 & 87.0 \\
Adversarial & 55.1 & 51.5 & 52.0 & 59.6 & 51.3 & 54.0 \\
\bottomrule
\end{tabular}
\caption{Per-category Judge Accuracy (\%) for the three robustness configurations under both judges. Baseline = OpenAI embedding + GPT-4o backbone; + DS bk = OpenAI embedding + DeepSeek-V3.2 backbone; + bge-m3 = bge-m3 embedding + GPT-4o backbone. The category-level profile is preserved across configurations (each category shifts by $\leq 4$~pp under either judge); bge-m3 improves Multi-hop by +3.8~pp under the GPT-4o judge.}
\label{tab:robustness_per_category}
\end{table*}

\onecolumn
\section{Prompt Templates}
\label{app:prompts}

This appendix lists the four LLM prompts used by \retmod{} and our evaluation pipeline. Placeholders are shown in \texttt{\{braces\}}; system messages and tier-specific variants are noted inline.

\begin{promptbox}{Prompt Template for Sufficiency Evaluation (\retmod{})}
\textbf{System:} You are a helpful assistant that evaluates information sufficiency.

\medskip
\textbf{User:}
\begin{verbatim}
Question: {question}

Retrieved Facts:
{facts}

[If TEMPORAL question, append:]
IMPORTANT: This is a TEMPORAL question asking for specific dates/times.
- EXACT_MATCH: Has precise date/time (e.g., "January 19, 2023")
- INFERRABLE: Has temporal clues that allow reasonable inference
  * "as of February 2023" -> event likely in February 2023
  * "after opening in January" -> subsequent events after January
- PARTIAL_MATCH: Has related but insufficient temporal information
- Vague terms like "recently" are PARTIAL, not EXACT.

Evaluate if these facts can answer the question:
1. EXACT_MATCH:    Can answer precisely?            (yes/no)
2. INFERRABLE:     Can reasonably infer the answer? (yes/no)
3. PARTIAL_MATCH:  Related but insufficient?        (yes/no)
4. CONFIDENCE:     0.0-1.0
5. MISSING:        what specific information is missing? (or "none")

Respond in EXACTLY this format:
EXACT: yes/no
INFERRABLE: yes/no
PARTIAL: yes/no
CONFIDENCE: 0.0-1.0
MISSING: <missing information or "none">
\end{verbatim}
\end{promptbox}

\newpage

\begin{promptbox}{Prompt Template for Query Refinement (\retmod{})}
\textbf{System:} You are a helpful assistant that refines search queries.

\medskip
\textbf{User:}
\begin{verbatim}
Original question:    {original_question}
Current search query: {current_query}
Missing information:  {missing_info}
Iteration:            {iteration}/{max_iterations}

Strategy: {strategy}

[Strategy is rule-based on (question_type, iteration):
  TEMPORAL,    iter 1: Focus on DATE, TIME, temporal keywords
                       (when, started, launched, opened).
  TEMPORAL,    iter 2: Search for specific DATE FORMATS and
                       temporal relations (as of, after, before).
  TEMPORAL,    iter 3+: Try broader temporal context: related
                        events, milestones, timeframes.
  NON-TEMPORAL,iter 1: Focus on specific keywords, entity names,
                       and key concepts.
  NON-TEMPORAL,iter 2: Try different angle: related events,
                       attributes, or contextual information.
  NON-TEMPORAL,iter 3+: Use synonyms, broader concepts, or
                        implied relationships.]

Generate an improved search query to find the missing information.
Keep it concise and focused. Return ONLY the query (no explanation).
\end{verbatim}
\end{promptbox}

\newpage

\begin{promptbox}{Prompt Template for Answer Generation (\retmod{})}
\textbf{System:} You are a helpful assistant that answers questions based on provided facts.

\medskip
\textbf{User:}
\begin{verbatim}
{tier_adaptive_instruction}
{reasoning_chain_context}     (if multi-hop reasoning was built)

Question: {question}

Relevant Facts:
{facts}

Answer (be concise and direct):
\end{verbatim}

\smallskip
\textbf{tier\_adaptive\_instruction} (selected by (\textsc{tier}, is\_temporal)):
\begin{verbatim}
TEMPORAL + EXACT:
  Answer with the PRECISE date/time. Use formats like
  "January 19, 2023" or "2023-01-19". Do NOT use vague
  terms like "around" or "approximately".

TEMPORAL + INFERRABLE:
  Make a careful inference from temporal clues. Use "in"
  or "around" if inferring timeframe; do NOT overuse
  "it is likely that" or "based on the facts".

NON-TEMPORAL + EXACT (or INFERRABLE, conf>=0.75):
  Answer directly and confidently. Do NOT hedge with
  "likely", "it seems", "probably", "based on the facts".

NON-TEMPORAL + INFERRABLE (or PARTIAL, conf>=0.50):
  Answer based on reasonable inference. Use "Based on
  the facts, [answer]" or state the answer with brief
  reasoning. Avoid overusing uncertainty markers.

LOW CONFIDENCE:
  Answer based on available facts. If key information is
  missing, state it concisely.
\end{verbatim}
\end{promptbox}

\newpage

\begin{promptbox}{Prompt Template for LLM-as-Judge Evaluation}
\begin{verbatim}
You are an impartial judge evaluating if an AI assistant's
answer is correct.

**Question**:     {question}
**Ground Truth**: {truth}
**Prediction**:   {prediction}

**Evaluation Criteria**:
1. The prediction must convey the same core information
   as the ground truth.
2. Different wording is acceptable if the meaning is
   preserved.
3. For dates: "May 7, 2023" and "7 May 2023" are
   equivalent.
4. If prediction says "I don't know" but ground truth
   exists, it is WRONG.
5. Partial answers that miss the key point are WRONG.

**Output Format**:
Return ONLY a JSON object:
{"score": 1 or 0, "reason": "Brief explanation"}
\end{verbatim}
\end{promptbox}

\end{document}